\documentclass[runningheads]{llncs}
\usepackage{float}
\usepackage[dvipdfmx]{graphicx}
\usepackage{color}
\usepackage{amsmath}
\newcommand{\Tag}[1]{{\it #1}}
\newcommand{\Font}[1]{{\tt #1}}

\usepackage{bbm}

\begin{document}

\title{Font Impression Estimation in the Wild}
%
%
\author{Kazuki Kitajima\and
Daichi Haraguchi\orcidID{0000-0002-3109-9053}\and\\
Seiichi Uchida\orcidID{0000-0001-8592-7566}}
\authorrunning{K. Kitajima et al.}

\institute{Kyushu University, Fukuoka, Japan\\
\email{uchida@ait.kyushu-u.ac.jp}}
\maketitle              
\begin{abstract}
This paper addresses the challenging task of estimating font impressions from real font images. We use a font dataset with annotation about font impressions and a convolutional neural network (CNN) framework for this task. 
However, impressions attached to individual fonts are often missing and noisy because of the subjective characteristic of font impression annotation. 
To realize stable impression estimation even with such a dataset, we propose an exemplar-based impression estimation approach, which relies on a strategy of ensembling impressions of exemplar fonts that are similar to the input image.
In addition, we train CNN with synthetic font images that mimic scanned word images so that CNN estimates impressions of font images in the wild.
We evaluate the basic performance of the proposed estimation method quantitatively and qualitatively. Then, we conduct a correlation analysis between book genres and font impressions on real book cover images; it is important to note that this analysis is only possible with our impression estimation method. The analysis reveals various trends in the correlation between them --- this fact supports a hypothesis that book cover designers carefully choose a font for a book cover considering the impression given by the font. 
\end{abstract}
%
\section{Introduction\label{sec:intro}}
%
\begin{figure}[t] 
\includegraphics[width=\textwidth]{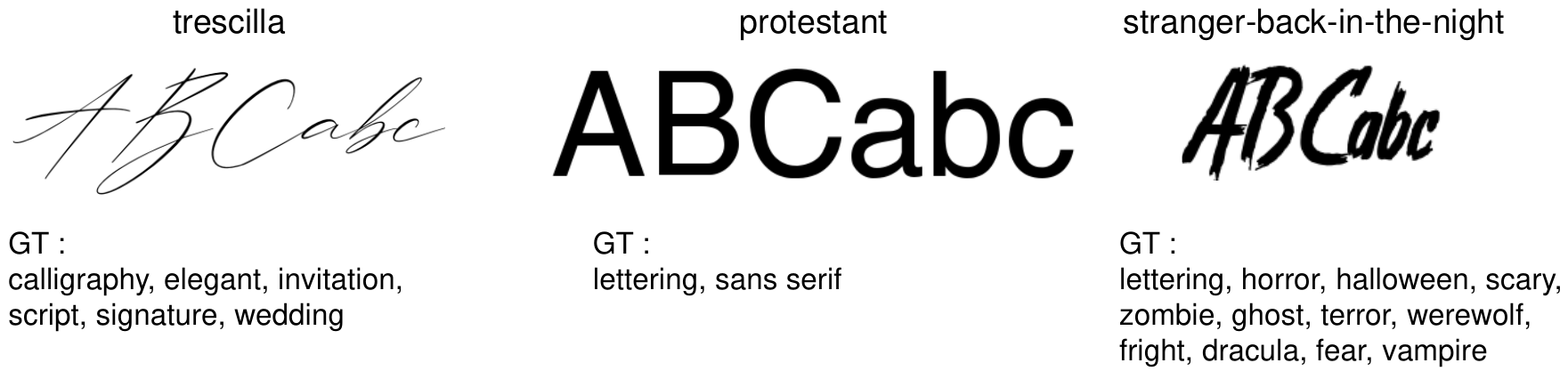} \\[-7mm]
\caption{Fonts and their impressions tags, provided by \url{1001freefonts.com}.}
\label{fig:font-example}
\end{figure} 
Each font has its own style and gives various impressions according to the style. 
Fig.~\ref{fig:font-example} shows examples of fonts and their impression tags. 
These examples are picked up from a website~\footnote{\url{https://www.1001freefonts.com/}}, where font designers freely attach these impression tags. 
It is noteworthy that each font has multiple and diverse impression tags. For example, in Fig.~\ref{fig:font-example}, the font {\tt trescilla} has diverse tags, such as \Tag{calligraphy}, \Tag{elegant}, \Tag{wedding}, etc. These impression tags are sometimes words describing font shapes directly (such as \Tag{Sans-serif}) rather than words representing abstract ideas, subjective feelings, and sensations. Throughout this paper, all of these tags are referred to as impression tags without discrimination, unless there is confusion.\footnote{In fact, discrimination of impression and non-impression tags is an interesting research topic by itself. The difference between them is very ambiguous. For example, the tag \Tag{bold} can be impression and non-impression. The tag \Tag{heavy} is rather an impression tag but can be a non-impression tag if it refers to a typographic term. These examples simply indicate that tags cannot be separated into impression and non-impression classes but are on a continuous scale, like data from regression or ranking tasks. The authors expect that studies like this paper will help to understand this continuous structure of impression tags.} \par
\begin{figure}[t] 
\centering
\includegraphics[width=0.8\textwidth]{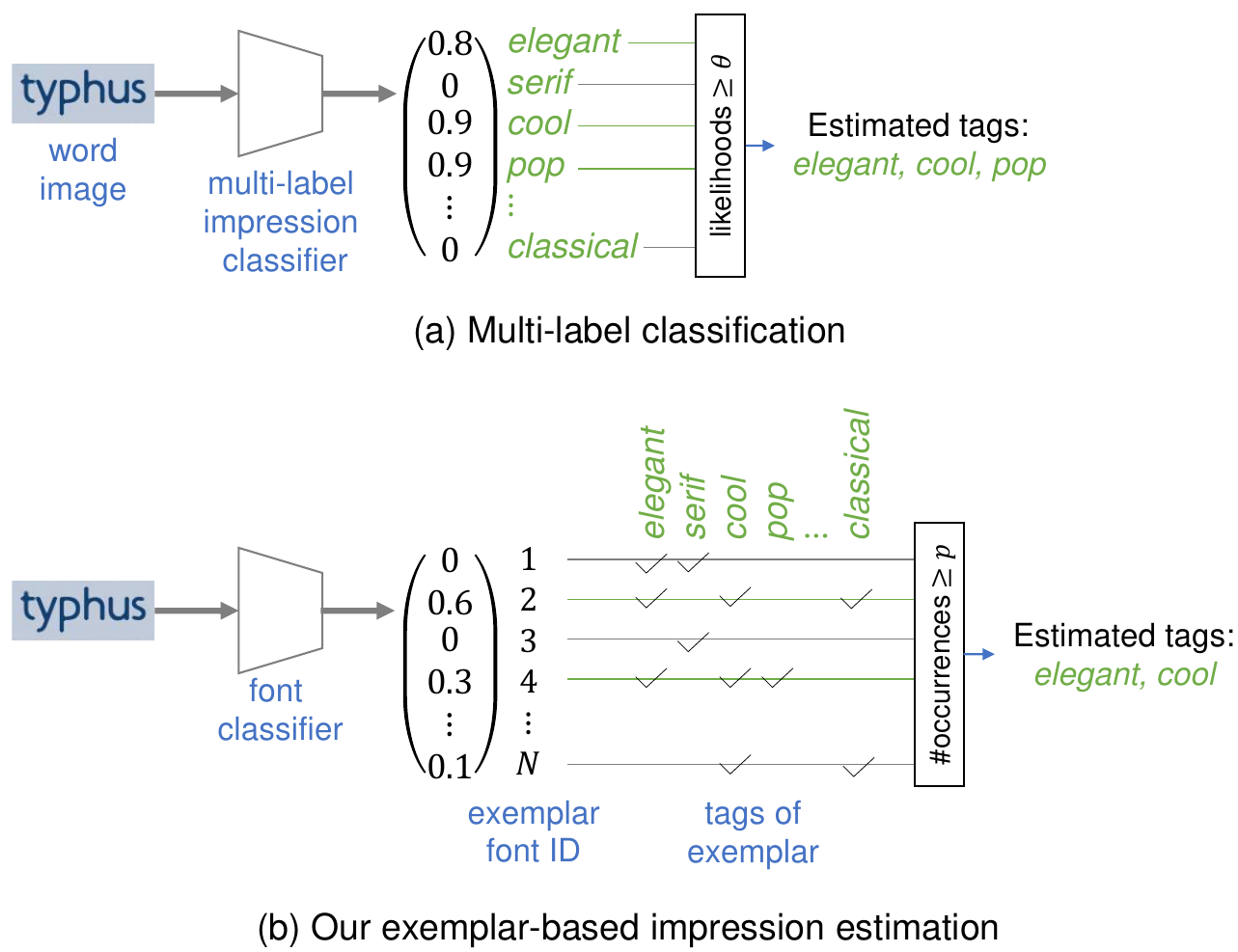}\\[-2mm]
\caption{Two approaches for the font impression estimation task. (a)~Multi-label classification, and  (b)~Our exemplar-based impression estimation. This is an example of the conditions $\theta = 0.8$, $\tilde{n} = 2$ and $p = 2$.} 
\label{fig:overview}
\end{figure} 
This paper tackles the task of estimating font impressions of texts {\em in the wild}. This task is meaningful in various aspects. For example, we can know what impression we get from a typographic work without questionnaires and subjective tests. Assume photographs capturing packages of cookies. If we can estimate what kinds of impressions are expressed by the texts on the packages, we can understand whether the package design fits the cookies. Such understanding is also useful for designing new cookie packages. More generally, it helps to know the trends of font impressions in various typographic designs.\par
Our task is not straightforward because our target is texts in scanned or photographed images. If we only deal with digital documents (like PDF),  it is easy to estimate the impressions of the fonts in them; this is because 
they include the list of fonts used in the document as metadata. The font impression tags of the text are readily obtained by searching the MyFonts website for the font name. However, text images in the wild do not have such metadata, and thus, we need to estimate font impression tags without knowing the font name.\par
One approach to estimating font impressions from a text image $\mathbf{x}$ is the multi-label image classification, as shown in  Fig.~\ref{fig:overview}~(a). This approach can be realized by a standard convolutional neural network (CNN). Since each font has multiple impressions (as we saw in Fig.~\ref{fig:font-example}), the CNN should be trained by supervision with $K$-dimensional multi-hot ground-truth vectors instead of one-hot vectors, where $K$ is the size of the impression tag vocabulary. \par
However, this strategy results in poor estimation performance due to the incompleteness of the ground-truth. More specifically, even when the $k$-th element of a $K$-dimensional ground-truth vector is zero, it often does not mean that the font does not show the $k$-th impression. 
The website \url{1001freefonts.com} distributes many fonts, and font creators {\em freely} attach the impression tags when uploading the fonts on the website. Consequently, even if a certain font has the $k$-th tag, a similar font might not have the tag. This is the so-called ``missing-label'' problem and disturbs multi-label classification significantly.
If we train a classifier with ground-truth vectors with missing labels, the classifier wrongly learns that the font does not show the $k$-th impression. It is inherently difficult to avoid the missing-label problem, especially for the impression labels. Even for font experts, it will be hard to prepare a unanimous ground-truth vector for a certain font by determining whether the font shows each of $K$ impressions.
\par
In this paper, we take another approach, exemplar-based impression estimation, instead of multi-label classification. Fig.~\ref{fig:overview}~(b) shows our approach. In this approach, we first prepare a $N$-class font classifier by a CNN trained to classify the input text image $\mathbf{x}$ into one of $N$ exemplar fonts, $\mathcal{R}=\{R_1, \ldots, R_N\}$. The classifier can give a similarity between the font of the input text and the $n$-th exemplar font as a probability of the $n$-th class, $P(R_n \mid \mathbf{x})$. Now, we assume that the impression tags of each exemplar font are available from a dataset. Then, we can expect that the font of the text in $\mathbf{x}$ will have the tags of the $n$-th font with the probability of $P(R_n \mid \mathbf{x})$. More simply, if $P(R_n \mid \mathbf{x})$ is high, we select the tags of $R_n$ as the font impressions of $\mathbf{x}$. If we have many exemplars whose probability is high, we accumulate their tags.\par
Although our exemplar-based approach is also simple, it achieves far better estimation accuracy than the multi-label classification approach of (a), as shown by the later experiments. In the exemplar-based approach, we do not use the impression tags in the training phase. This means that the trained model is not affected by missing labels. Moreover, even if a tag is missed in a certain exemplar $R_n$, we can expect that the tag will be found in similar exemplars and, consequently, selected as one of the estimated impression tags.
\par
After evaluating the exemplar-based approach with synthetic images, we apply it to real book cover images. On book cover images, many texts are printed as book titles, authors, publishers, etc., and they are printed in different font styles with different impressions. Former studies~\cite{yasukochi2023analyzing,shinahara2019serif} have revealed correlations between book genres and font styles (i.e., font shapes) on book covers. However, they did not provide any results about font impressions. Using our exemplar-based impression estimation, we can directly analyze the correlation between book genres and impressions. This new analysis will give perceptual correlations between book genres and impression tags, and these correlations are more directly understandable than the past analyses. \par
The main contributions of this paper are summarized as follows:
\begin{itemize}
    \item To the best of the authors' knowledge, this is the first attempt to estimate impressions from not born-digital font images but word images in the wild.
    \item The proposed exemplar-based impression estimation method shows high robustness to both missing impressions and noisy impressions, which are inevitable when dealing with subjective impressions.
    \item As an application task, we analyze the correlation between book genres and the impression of fonts on the book cover. 
\end{itemize}
\color{black}
\section{Related Work}
\subsection{Impressions of Fonts} 
The relationship analysis between fonts and their impression has been considered as an important research topic from both practical and scientific viewpoints and thus has a long history since the 1920s~\cite{davis1933determinants,poffenberger1923study}.
In these early studies, the target fonts are very limited (just a dozen of fonts).
In 2004, Henderson~\textit{et al.}\cite{henderson2004impression} analyzed more than 200 fonts and their impressions to create font usage guidelines for marketing.
In a more recent study, Choi and Aizawa~\cite{choi2019emotype}  used 100 fonts to analyze the effect of fonts on emotions in mobile text messages.\par
Font datasets with impression tags have been used in various font-related tasks, especially font retrieval (or recommendation) and font generation. In font retrieval, O'Donovan \textit{et al.}~\cite{o2014exploratory} are pioneers of a tag-based (i.e., impression-based) font retrieval task. They collected tags of fonts through crowd-sourcing and proposed a tag-based font retrieval interface. Following this study, several tag-based font retrieval approaches have been proposed~\cite{choi2019assist,kulahcioglu2020fonts}.\par
In font generation, Attribute2font~\cite{wang2020attribute2font} is the pioneering attempt to generate fonts with attribute tags (i.e., impression tags). Following this attempt, several impression-conditioned font generation models have been proposed~\cite{matsuda2022font,matsuda2021impressions2font,kang2022shared}. For example, Matsuda~\textit{et al.}~\cite{matsuda2021impressions2font} proposed Impressions2font that can generate font images from impressions considering missing properties of impression tags.
Kang~\textit{et al.}~\cite{kang2022shared} created the shared latent space between word embeddings of impression and font image features. Through the shared latent space, they realize both font generation and font retrieval from the impression by a single neural network.\par
We can also find several papers that tackle the font impression estimation from images~\cite{choi2019assist,ueda2022font,ueda2021parts}. 
To the best of the authors' knowledge, these previous attempts cannot deal with text images in the wild because they are designed to estimate the impressions of ``born-digital'' font images directly created from true-type font files.
For example, the part-based approach of \cite{ueda2022font,ueda2021parts} extracts keypoints like SIFT from a born-digital single-letter font image and uses them for the estimation; therefore, it does not expect many spurious keypoints from the background region of the text image in the wild. Moreover, none of them explicitly addresses the existence of missing impressions and noisy impressions simultaneously.\par
%
\subsection{Font Usage Analysis\label{sec:review-genre}}
As noted in Section~\ref{sec:intro}, different fonts have different impressions, and therefore, typographic experts carefully choose fonts for their typographic works, considering the appropriateness of font styles to the typographic context of the works. This means if we analyze the relationship between font styles and contexts from the existing typographic works, we can know the experts' knowledge and then utilize it in typographic works in the future. Depending on what we consider as the typographic context of style, we can consider various variations of analysis tasks~\cite{wayne2019kept,choi2016typeface,kulahcioglu2018fontlex,yasukochi2023analyzing,shirani2020let,kulahcioglu2019paralinguistic}. For example, 
Shirani~\textit{et al.}~\cite{shirani2020let} analyzed the relationship between the font styles and the verbal context printed in the font style; they performed a user study to choose the most appropriate font style for a certain phrase, such as ``My first surf day.''  Kulahcioglu~\textit{et al.}~\cite{kulahcioglu2019paralinguistic} pick up emotions as the context; according to user studies on synthetic word clouds printed with various font styles, they analyze the relationship between font styles and their emotional impressions, such as \Tag{calm} and \Tag{serious}. \par
There are several attempts to analyze the font styles on book covers~\cite{shinahara2019serif,yasukochi2023analyzing}. Those attempts consider book genres as the context of the font style. They revealed correlations between styles and genres; for example, sans-serif fonts are often used for engineering books, not religion-related ones. A limitation of these attempts is that they represent font styles by numerical vectors derived from font shapes, and thus, the interpretation of the correlation is not straightforward. In contrast, this study analyzes the correlation by representing font styles as impression tags. As a result, the relationship between style and genre can be {\em described by words} about impressions and understood more intuitively.\par
\section{Font Dataset\label{sec:dataset}}
We collected 4,000 fonts and their impression tags from \url{1001freefonts.com}. 
As shown in Fig.~\ref{fig:font-example}, font creators can attach multiple impression tags to each font on \url{1001freefonts.com} when they upload their fonts. The average number of tags attached to a font is about 4.\par
\begin{figure}[t] 
\includegraphics[width=\textwidth]{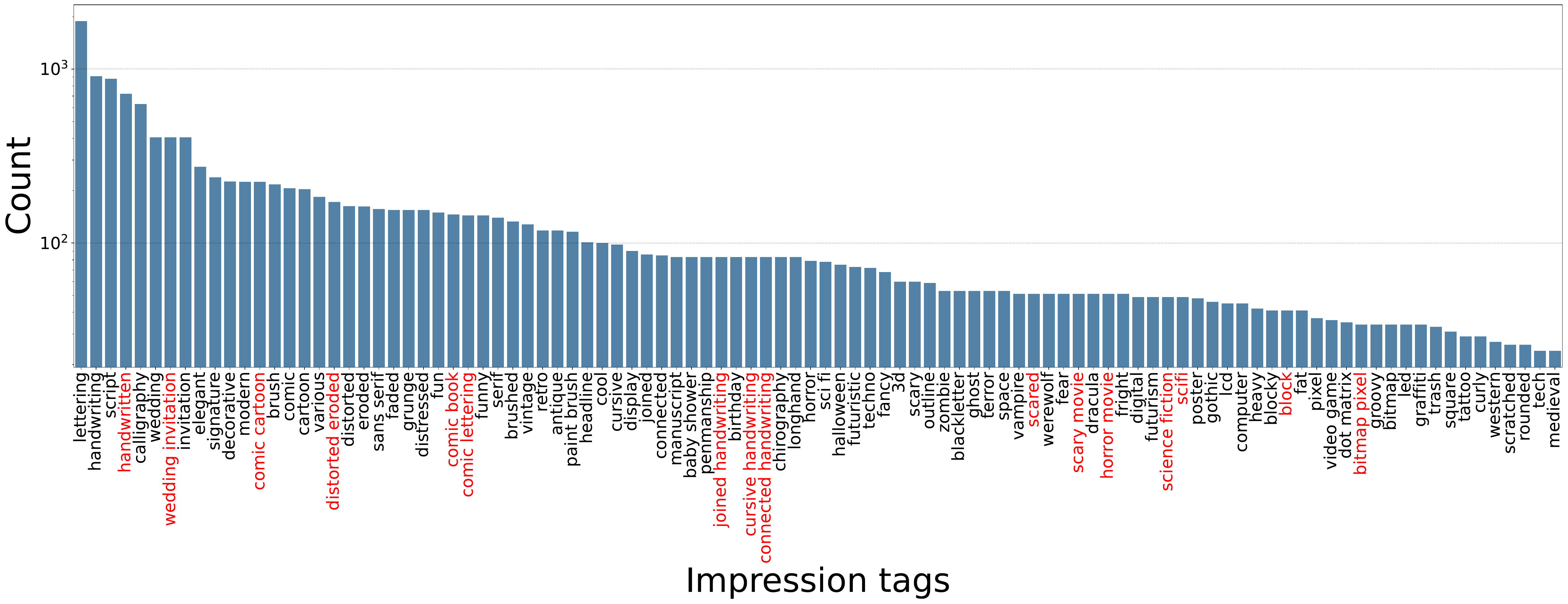}\\[-3mm]
\caption{The number of fonts with the 100 most frequent impression tags. Note that the vertical axis is logarithmic and the 16 tags in red are ignored due to the existence of near-identical tags.}
\label{fig:tag-histogram}
\end{figure}

The dataset has 589 different impression tags. Fig.~\ref{fig:tag-histogram} shows the 100 most frequent impression tags and their occurrences in the dataset.
We can see the occurrences of tags are quite imbalanced. The first tag \Tag{lettering} has more than 2500 occurrences, whereas the 100th tag \Tag{medieval} has just 24. We thus ignore the tags with lower occurrences than the 100th tag because they do not have enough occurrences. Among the 100 tags, several are merged when two have a minor spelling difference. 
For example, \Tag{scifi} and \Tag{science fiction} are merged into \Tag{sci fi}. 
Second, several compounded tags are ignored due to duplication;  for example, the tag \Tag{comic cartoon} is comprised of two top-100 tags \Tag{cartoon} and \Tag{comic} and thus ignored. (In this case, a font with \Tag{comic cartoon} is treated as a font with \Tag{cartoon} and \Tag{comic}.) After removing these duplicated tags, we used $K=84$ tags in our experiments. \par
It should be emphasized that impression tags in the dataset are still noisy and incomplete (even after removing less frequent and duplicated tags). Font creators freely attach impression tags according to their thoughts. In addition, they do not need to attach all possible tags; therefore, there are many missing tags. For example, very thick fonts are often not tagged as \Tag{heavy}. Consequently, the absence of an impression tag does {\em not} indicate that the font does not show the impression, and a classifier trained by a multi-hot vector specified by the presence or absence of individual impressions is not appropriate.\par
\section{Exemplar-based Impression Estimation}
%
We assume an exemplar font set $\mathcal{R}=\{R_1, \ldots, R_n, \ldots, R_N\}$. For each exemplar font $R_n$, a set of impression tags (such as \Tag{calligraphy}, \Tag{elegant}, $\ldots$, \Tag{wedding} of \Font{trescilla} in Fig.~\ref{fig:font-example}) is attached as $T_n$. As emphasized so far, the impression tags may be incomplete; there might be some missing tags in $T_n$ that are still appropriate for $R_n$.\par 
As shown in Fig.~\ref{fig:overview}, our exemplar-based font impression estimation uses a CNN trained for an $N$-class font classification task. Specifically, word images, each of which is printed with a certain exemplar font in $\mathcal{R}$, are used for training the CNN. (As detailed in Section~\ref{sec:SynthTiger}, the word images are generated by SynthTiger~\cite{yim2021synthtiger} so that the images look like the texts in the wild.) The output of the CNN is an $N$-dimensional vector $\mathbf{y}=(y_1, \ldots, y_n, \ldots, y_N)$, where $y_n$ shows the likelihood that the input image $\mathbf{x}$ is printed with $R_n$.\par   
Using the classification result $\mathbf{y}$, we finally estimate the font impressions of $\mathbf{x}$ by an ensemble strategy. More specifically, we aggregate all tags of the $\tilde{n}$ exemplar fonts with the $\tilde{n}$ highest $y_n$ values. This aggregation operation relaxes the missing-label problem; the $\tilde{n}$ exemplar fonts will compensate for each other's missing tags. Then, we count how often each tag occurs in the aggregation result. Finally, we select the impression tags with more than $p$ occurrences as the font impressions of $\mathbf{x}$. This selection with the threshold $p$ is necessary to remove noisy labels.
\section{Impression Estimation Experiment on Synthetic Word Images}
%
\begin{figure}[t]
 \begin{center}
\includegraphics[width=\textwidth]{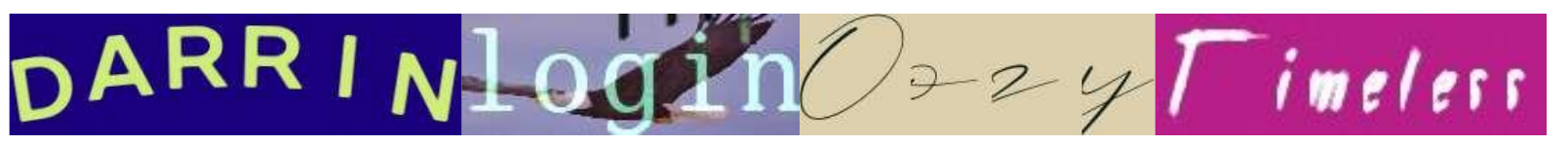}\\[-4mm]
\caption{Examples of the synthetic word images for training the font classifier. Images
are generated by SynthTiger~\cite{yim2021synthtiger}.}  \label{fig:synthtiger_examples}
\end{center} 
\end{figure}
%
\subsection{Synthetic Word Images\label{sec:SynthTiger}}
SynthTiger~\cite{yim2021synthtiger} is used for preparing synthetic word images. SynthTiger generates a word image with a font on a background image. Examples of synthetic word images are shown in Fig.~\ref{fig:synthtiger_examples}. The $4,000$ fonts in Section~\ref{sec:dataset} are splitted into $3,000$, $300$, and $700$, for training, validation, and testing.  
This means we use $N=3,000$ exemplar fonts.
The words are randomly chosen from the $88,173$ words in a word corpus, MJsynth~\cite{Jaderberg14c}. The background images are picked up from the SynthTiger GitHub. Although the size of each generated word image is variable, the image is resized to $64\times128$ as the input image to CNN. When resizing, zero padding is performed to keep the original aspect ratio of the word in the input image.
\par %
We generate $3,000\times 900$ images by SynthTiger for training the 3,000-class font classification model, where $900$ is the number of words. The detail of the training procedure is given in Section~\ref{sec:details}.
We also generate $300 \times 100$ images and $700\times 100$ images as the validation and test sets, where $100$ is the number of words. The validation set is used to fix hyperparameters.
\par %
\subsection{Font Classification Model and Its Training\label{sec:details}}
We used ResNet-18~\cite{he2016deep} as the font classification model.
The $3,000\times 900$ synthetic word images are further decomposed into 
$3,000\times 800$ images for training and $3,000\times 100$ images for validation. 
The training process follows a standard scheme: cross-entropy loss, SGD, and a batch size of $128$. The learning rate is set to an initial value of $0.1$ and gradually decreases to $0.01$ as the number of epochs increases, following a cosine curve. The model with the best validation accuracy during $200$ epochs is selected as the trained model.\par
%
\subsection{Comparative Models}   
%
We prepared three multi-label classification models, which can directly estimate the impression tags of input fonts. All models are based on ResNet-18 and trained with two loss functions: mean squared error (MSE) and binary cross entropy (BCE). For BCE, two versions were examined, one with and one without class weights. The class weight is introduced to balance the tag occurrence. For the $k$th impression, the class weight is given as $\frac{M-M_k}{M_k}$, where $M$ is the number of all training images (i.e., $3,000\times 800$) and $M_k$ is the number of training images with the $k$-th tag. We trained all three models in the same training setting as ours.
%
\subsection{Evaluation Metrics and Hyperparameters}
%
We evaluated the impression estimation performance using recall, precision, and F1-score. Considering the heavy class imbalance situation, each metric is calculated by the macro criterion, where recall, precision, and F1-score are calculated at each tag and then averaged in all tags.\par
We used the F1-score on the validation set to optimize the hyperparameters, 
$\tilde{n}$, $p$, and $\theta$. As the optimization results, $p=3$ and $\tilde{n}=11$. The thresholds $\theta$ for the multi-label classification models by MSE, BCE without class weights, and BCE with class weights were $0.1$, $0.1$, and $0.7$, respectively.
\begin{table}[t]
\caption{Evaluation of the estimated impressions on synthetic word images.}
\label{comparison}
\centering
\begin{tabular}{l|r|r|r}\hline
& Recall $\uparrow$ & Precision  $\uparrow$ & F1 score $\uparrow$ \\ \hline
Ours                &$0.3507$ & $\mathbf{0.2870}$ & $\mathbf{0.2913}$         \\ 
Multi-label CNN (MSE)                  & $0.1500$ & $0.0401$ & $0.0583$         \\
Multi-label CNN (BCE w/o class weights)                        & $0.3333$ & $0.1588$ & $0.2001$         \\
Multi-label CNN (BCE w class weights)                         & $\mathbf{0.4583}$ & $0.1141$ & $0.1549$         \\ \hline
\end{tabular}
\end{table}
\begin{figure}[t]
\centering
   \includegraphics[width=0.8\textwidth]{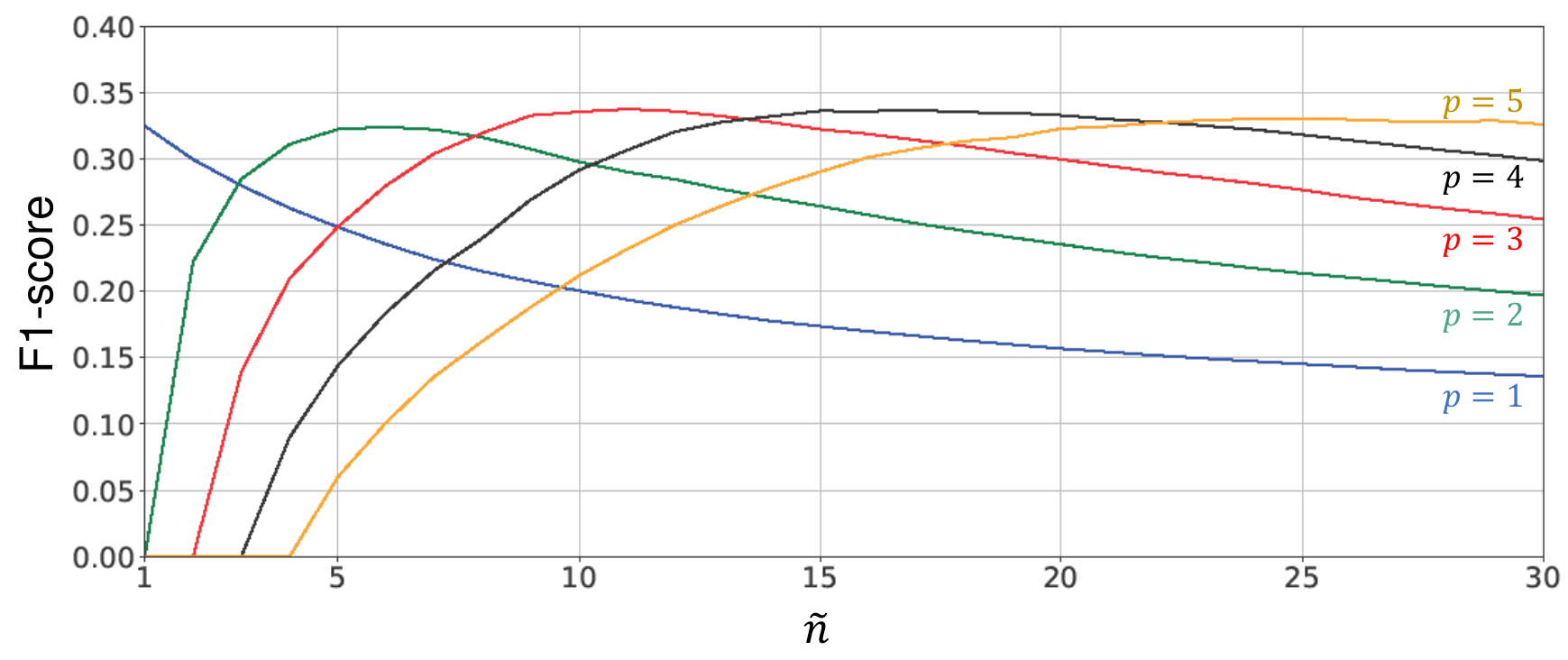}\\[-4mm]
\caption{F1 scores under the different hyperparameter values, $\tilde{n}$ and $p$.}
\label{fig:f1_measure}
\end{figure}
%
\subsection{Quantitative Evaluation Results}   
%
Table~\ref{comparison} shows the quantitative evaluation results. As summarized by the F1 score, ours achieved a better impression accuracy than the multi-label CNNs regardless of their loss functions and introduction of class weights. Ours achieves the highest precision and second-highest recall because ours uses the ensemble strategy to remove noisy labels (for higher precision) and compensate for missing labels (for higher recall).
Note that introducing class weights degrades overall performance (F1) drastically for the multi-label CNN. This is because it boosts noisy (less frequent) tags and worsens the precision.
\par
Fig.~\ref{fig:f1_measure} shows how the F1 score of the our approach changes by the hyperparameters, $\tilde{n}$ and $p$. This plot indicates the effectiveness of our ensemble strategy with these hyperparameters. For example, when we do not discard less frequent tags from the aggregated tags by setting $p=1$, F1 scores degrade according to $\tilde{n}$; more tags from more exemplar fonts contain more noisy labels. By using $p>1$, we can rely on more exemplars (i.e., more $\tilde{n}$) for better impression estimation.
\par
%
\subsection{Qualitative Evaluation Results}  
\begin{figure}[t] 
\includegraphics[width=0.98\textwidth]{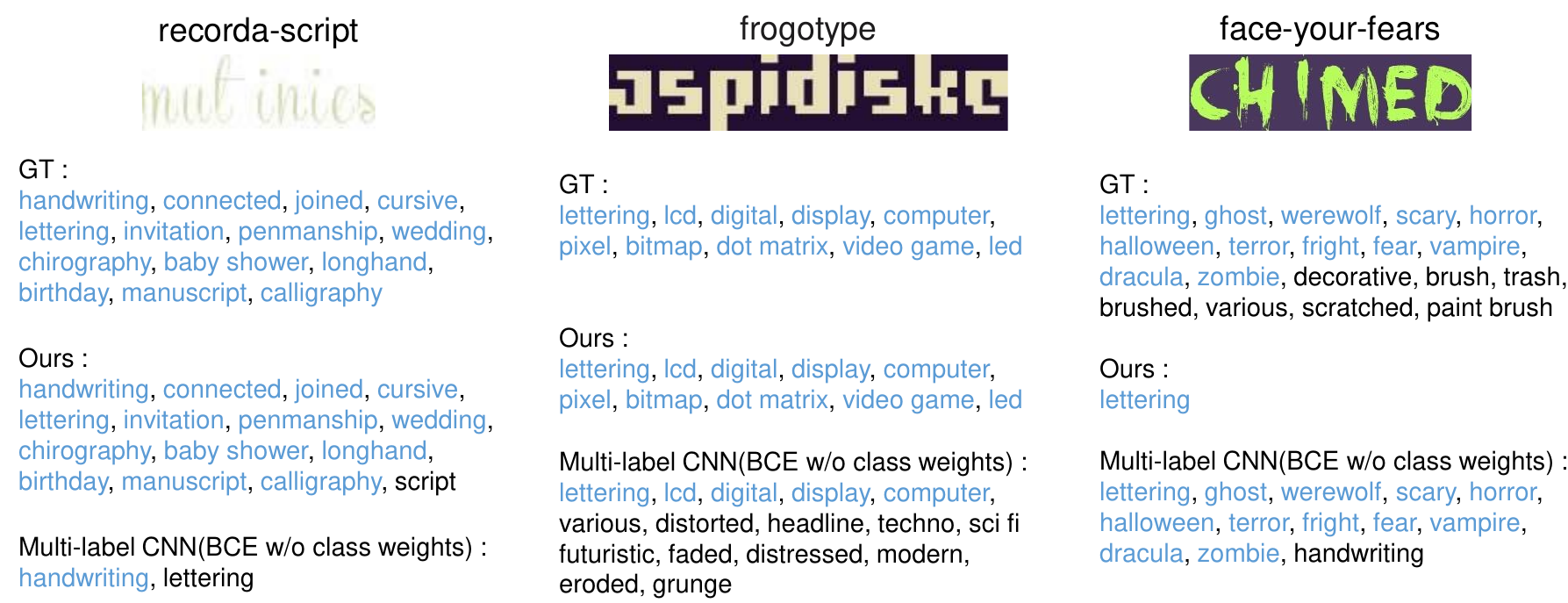}\\[-4mm]
\caption{Examples of synthetic font images and estimated impressions. The blue tags are true positives.} \label{fig:font-impressions}
\bigskip
\bigskip
\includegraphics[width=0.98\textwidth]{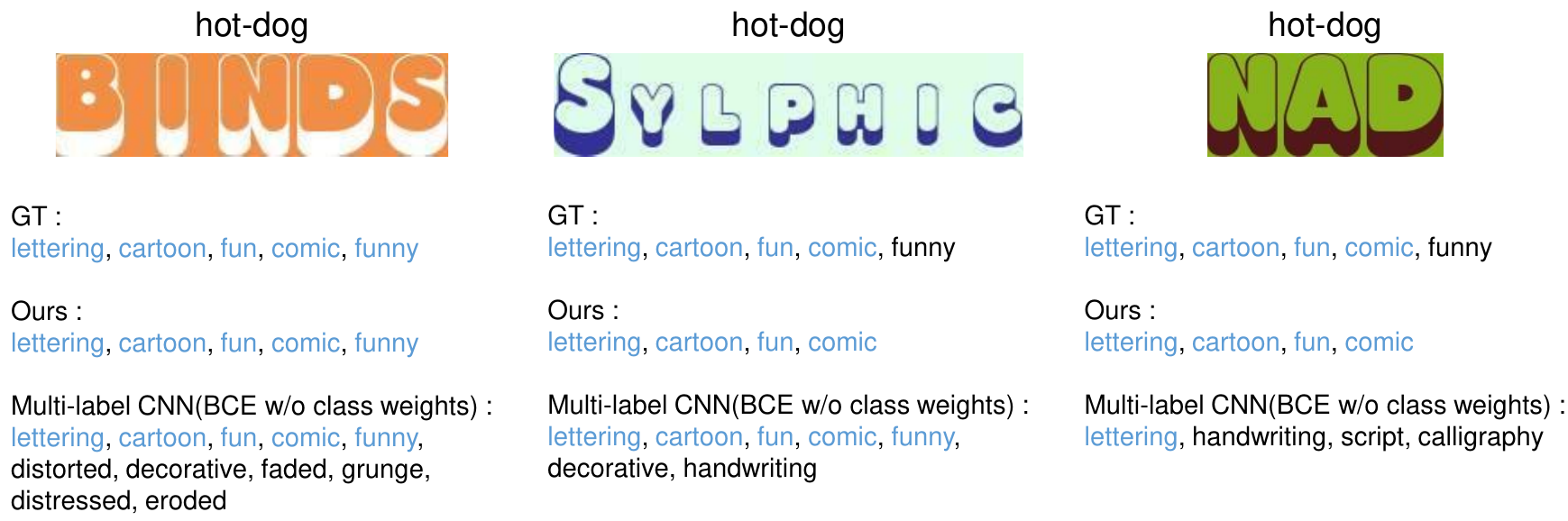}\\[-4mm]
\caption{Stability of impression estimation for word images of the same font. The blue tags are true positives.} \label{fig:font-impressions-consistency}
\end{figure}

%
Fig.~\ref{fig:font-impressions} shows the impression estimation results by the our approach and the multi-label CNN (with the BCE loss and without class weights) on several synthetic images by SynthTiger. Each image is printed with a font whose impression tags are known as ground-truth (GT). For the fonts \Font{recorda-script} and \Font{frogotype}, the our approach gives (almost) perfect estimation results. In contrast, the multi-label CNN gives rather irrelevant impressions and many missing impressions. The rightmost shows a failure case, where only a single impression \Tag{lettering} was given for the font \Font{feace-your-fears}. This case originally had more (correct) tags, such as \Tag{ghost}, \Tag{horror}, and \Tag{scary}, but these tags were discarded by the threshold $p$.\par
Fig.~\ref{fig:font-impressions-consistency} shows the impression estimation results on word images printed in the same font (called \Font{hot-dog}). If an estimator is stable enough, all of them should have the same impression estimation result. Our method successfully gives almost the same impressions to the three images.
In contrast, the multi-label classification method shows unstable estimation results and even inappropriate impressions, such as \Tag{handwriting}, \Tag{script}, and \Tag{calligraphy}. \par
\section{Application: Correlation Analysis Between Book Genres and Font Impression on Book Covers}
\subsection{Purpose}  
%
A certain font will be chosen by a typographer when the impression of the font matches the context of their target work. For example, fonts with funny impressions can be candidates for a comedy movie poster. For a signboard of a sophisticated restaurant, fonts with elegant impressions might be chosen. The correlation between font impressions and their contexts found in many actual use cases is useful to evaluate how a chosen font is appropriate for a typographic work. Moreover, we can use the correlation when we manually or automatically generate new typographic work. \par
In this section, as an example of actual font use cases, we use about 200,000 book cover images. Professional typographers generally design book covers and carefully choose fonts for a cover while considering how the fonts are appropriate to the book's content. Therefore, we expect book covers to show a stronger correlation between font impressions and contexts (i.e., book contents) than other use cases. Here, we employ the {\em genre} of the book as a simple but clear index for representing the content of a book. Consequently, we aim to find the correlation between font impressions and book genres using many book cover images. Different from digital documents, book cover images are just raster images. Therefore, we need to detect word regions from them and then estimate their impressions using our exemplar-based impression estimation method.\par
As noted in Section~\ref{sec:review-genre}, previous studies~\cite{shinahara2019serif,yasukochi2023analyzing} revealed correlations between font {\em styles} and book genres. However, the font style is often represented as a non-intuitive numerical feature vector~\cite{yasukochi2023analyzing}. Therefore, even though it is possible to evaluate the correlations quantitatively, it is difficult to describe them. 
In contrast, we use impression words instead of style vectors 
and more intuitively describe the correlation between impressions and book genres. \par
%
\subsection{Book Cover Dataset and Impression Estimation}  
\begin{figure}[t] 
\includegraphics[width=\textwidth]{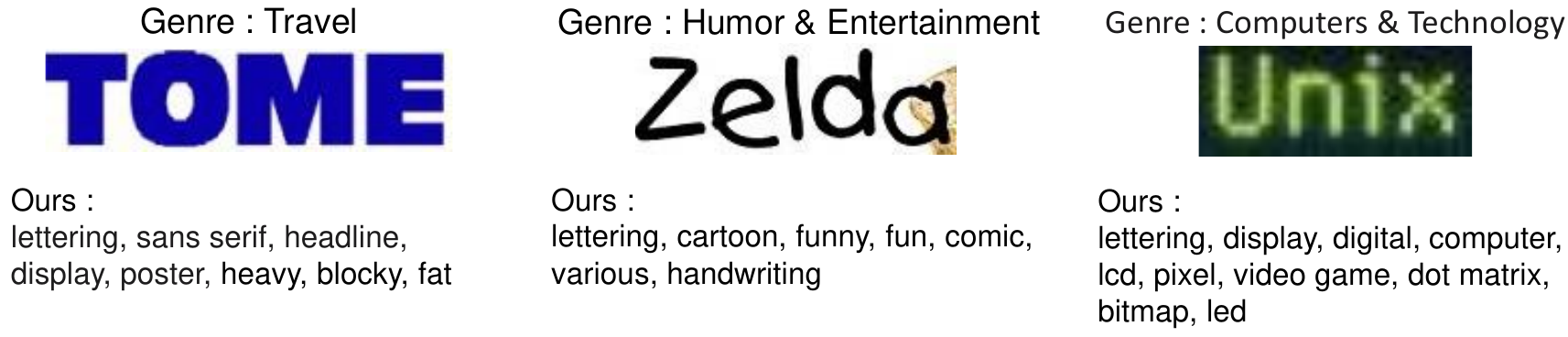}\\[-3mm]
\caption{Text images on book covers and their estimated impressions.\label{fig:bookcover-impression1}}
\end{figure}

%
We used the book cover dataset~\cite{iwana2016judging} with $207,572$ images from \url{Amazon.com}. For each book cover image, its title and one of 32 pre-defined genres are attached as metadata. Following~\cite{yasukochi2023analyzing}, word regions in each book cover image are extracted by a standard scene text detector, CRAFT~\cite{baek2019character}, and then each region is recognized into a word by an OCR model~\cite{baek2019wrong}. As post-processing, we discard regions showing words not found in the book title and regions with less than 15 pixels in height. Each word image is finally resized to $64\times 128$ by the same procedure of Section~\ref{sec:SynthTiger}. Fig.~\ref{fig:bookcover-impression1} shows three examples of word images from book cover images. The hyperparameters $k$ and $\tilde{n}$ were set at the same value as the impression estimation experiment.
\par
Fig.~\ref{fig:bookcover-impression1} shows the impression estimation results for several words extracted from book cover images. The results indicate that the our approach can estimate reasonable impressions. For example, visually-valid impressions, such as \Tag{sans serif} and \Tag{heavy}, are estimated for ``TOME.''
More interestingly, for ``Zelda'' from a book in the Humor \& Entertainment genre, the our approach estimates impressions, such as \Tag{fun} and \Tag{funny}, which are semantically consistent with the genre. 
\par
\begin{figure}[p]
\centering
\includegraphics[height=\textheight]{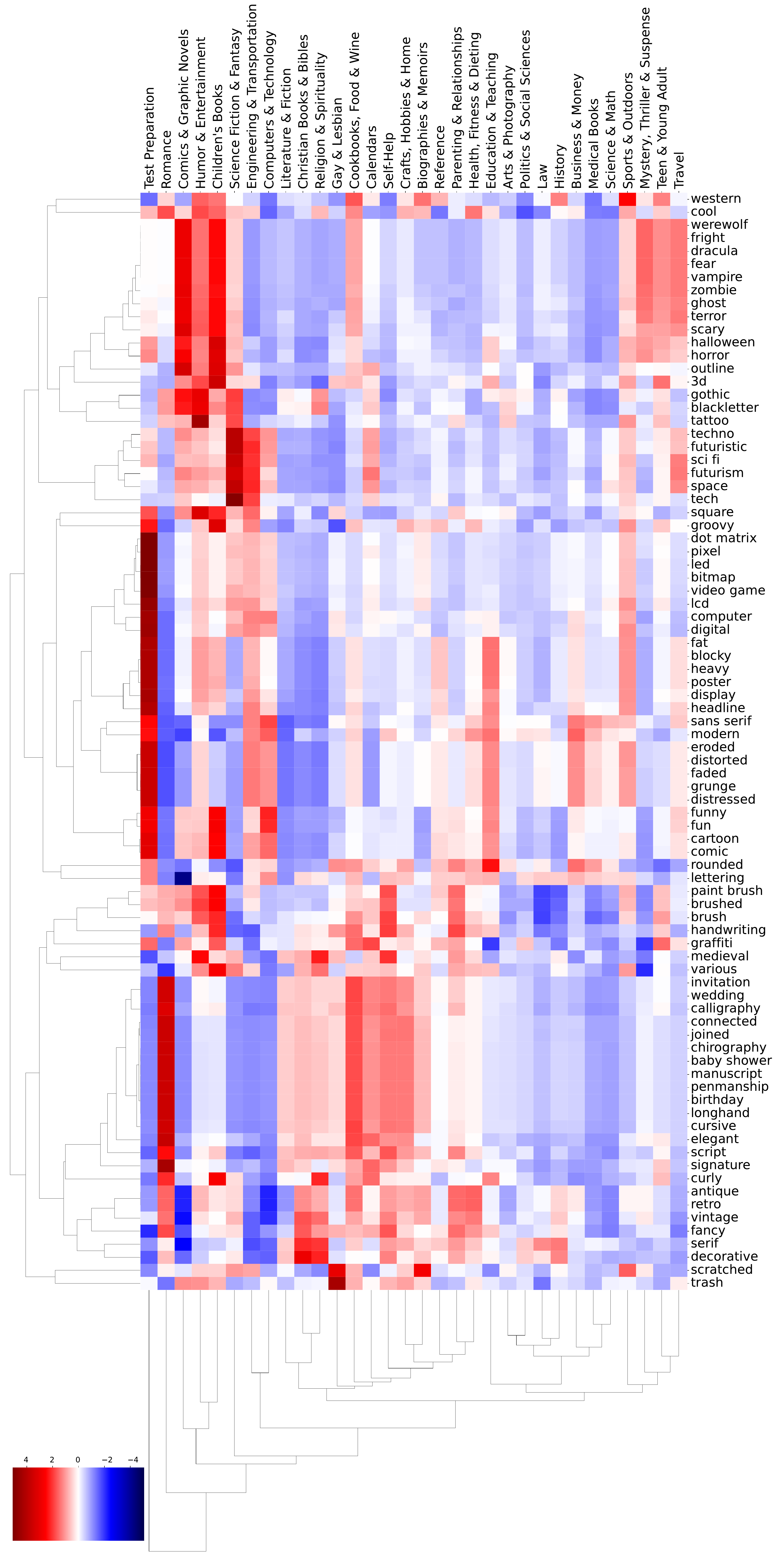}
\caption{The heatmap showing the correlation between book genres and impressions. Red (blue) means that this genre is more (less) frequent for the tag.} \label{fig:correlation-matrix}
\end{figure} 
\begin{figure}[t] 
\includegraphics[width=\textwidth]{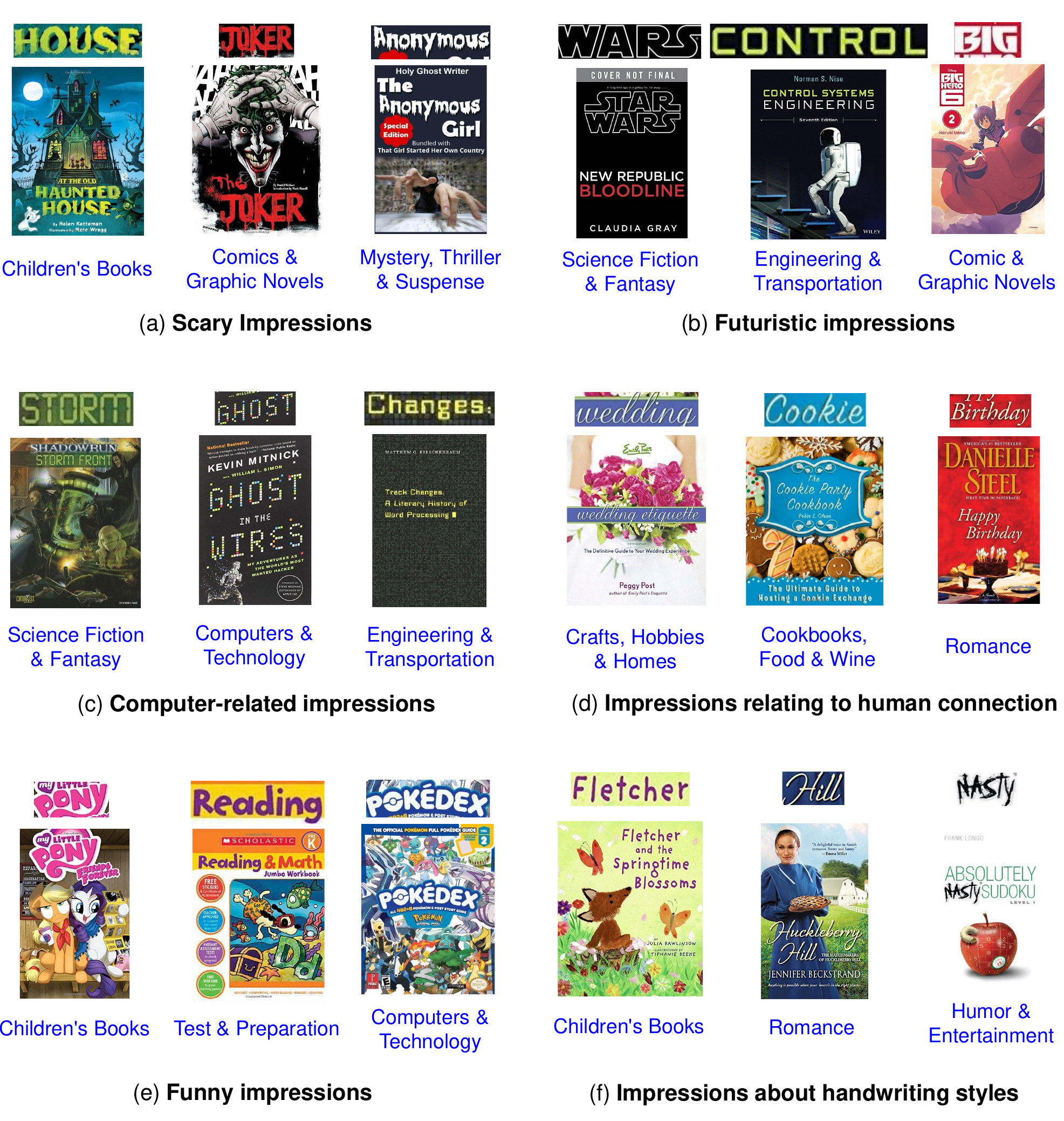}
\caption{Book cover images and their correlated impressions. Texts in blue are genres.} \label{fig:bookcover-impression2}
\end{figure} 
%
\subsection{Correlation Analysis between Book Genres and Impressions}  
%
The heatmap of Fig.~\ref{fig:correlation-matrix} shows the correlation between impressions and genres for all $207,572$ book images. 
To minimize the effect of title lengths (i.e., the number of words for each title), 
the impressions from title words are accumulated by the OR operation for each book. 
Then, we performed a two-step normalization: First, the occurrences of 84 impressions in each genre are normalized to eliminate the effect of differences in the number of books per genre. Second, the (normalized) occurrences  
in each tag are further normalized to mean 0 and variance 1. This two-step normalization allows us to observe the genre ratio of each impression, in other words, how each impression is distributed among the 32 genres. Finally, the rows and columns of the heatmap are sorted by biclustering so that similar rows 
are closer to each other and, simultaneously, similar columns are closer to each other. \par
The main findings from the heatmap are summarized as follows:
\begin{itemize}
\item First of all, this heatmap is not a flat white. If an impression tag appears in all genres with equal occurrence, this heatmap becomes white. This heatmap, however, is not white, and thus, there are various correlations between impressions and genres. 
\item There are several large red clusters, each of which shows a strong positive correlation between certain types of genres and impressions. 
\begin{itemize}
    \item Scary impressions such as \Tag{fear}, \Tag{horror}, and \Tag{ghost}, mainly occur in the entertainment genres, such as Comics \& Graphic Novels, Children's Books, and Mystery, Thriller \& Suspense. (Example: Fig.~\ref{fig:bookcover-impression2}(a))
    \item Futuristic impressions such as \Tag{futuristic}, \Tag{space} and \Tag{techno} occur in the scientific genres, such as Science Fictions \& Fantasy, Engineering \& Transportation. (Example: Fig.~\ref{fig:bookcover-impression2}(b)) Note that computer-related impressions, such as \Tag{led}, \Tag{bitmap} also show similar occurrence patterns, although they focus more on Computers\& Technology. (Example: Fig.~\ref{fig:bookcover-impression2}(c))
    \item Futuristic impressions are also found in the genres where the scary impressions are used (such as Comics\& Graphic Novels) --- however, there are also differences. For example, futuristic impressions are rather avoided in Mystery, Thriller \& Suspense, and scary impressions are avoided in Engineering \& Transportation.
    \item Impressions relating to human connections, such as \Tag{wedding}, \Tag{birthday} and \Tag{invitation}, occur in the home-oriented genres, such as Cookbooks, Foods, \& Wine and Crafts, Hobbies, \& Home. These impressions also occur in Romance and Self-Help. (Example: Fig.~\ref{fig:bookcover-impression2}(d))
    \item Funny impressions such as \Tag{fun}, \Tag{funny}, \Tag{comic}, \Tag{cartoon} occur in the Children's books. (Example: Fig.~\ref{fig:bookcover-impression2}(e))
\end{itemize}
\item There are several large blue clusters showing negative correlations.
\begin{itemize}
    \item For example, impressions about handwriting styles are avoided in society-oriented genres, such as Law and History. and also in the engineering genres. 
    \item Futuristic, funny, and computer-oriented impressions are avoided in literature, religion, and romance books. (In contrast, these impressions are welcome by Test Preparation.)
\end{itemize}
\item Other miscellaneous correlations are as follows.
\begin{itemize}
    \item Impression tags about handwriting styles show interesting trends.
     Artistic writing styles, such as \Tag{calligraphy}, \Tag{cursive}, and \Tag{penmanship}, show their main focus on the home-oriented genres, similar to the impressions about human connections. However, other writing styles, such as \Tag{brush}, \Tag{graffiti}, and \Tag{handwriting}, have more focus on Children's Books and Humor \& Entertainment. (Example: Fig.~\ref{fig:bookcover-impression2}(f))
     \item Classical impressions, such as \Tag{retro}, are used in various genres; However, this does not apply to the engineering genres.
     \item Scary impressions occur in Travel and Cookbooks, Foods, \& Wine, mainly because of travel-guide books to scary places and Halloween cookbooks. 
    \item Funny impressions are also used for the books in Education \& Teaching. Although it seems a bit mismatched, educational books for children use them.
\end{itemize}
\end{itemize}
\par
As summarized above, there are many correlations between genres and impressions. This fact proves that design experts of book covers are (implicitly or explicitly) considering the correlation between font impressions and book contents. Some of these correlations seem somewhat obvious to us according to our experiences --- however, the worth of this analysis is that we {\em objectively and quantitatively} prove the correlations solely based on data. (We also find several unexpected correlations --- impressions about writing styles are not monolithic.) 
At the same time, the fact that several obvious correlations are found (as expected) indicates our font impression estimation method works on text images in the wild and, therefore, is applicable to different impression analysis tasks.\par
\section{Conclusion, Limitation, and Future Work}
%
The main contributions of this paper are the proposal of a simple but accurate font impression estimation method and its application to actual typographic images, i.e., title word images in book covers. Our estimation method relies on the exemplar fonts; given an input word image, our method first selects several exemplar fonts similar to the input by a CNN and then ensemble the impressions attached to the selected exemplars. This ensemble strategy shows robustness to the practical problem that font impressions are often noisy and incomplete (i.e., missing). The application result to title word images reveals not only that our method works well on real word images but also that there are several strong correlations between font impressions of the title words and book genres. For example, fonts with \Tag{scary} impressions are often employed in the genres of cartoons, children's books, and mystery. Fonts with \Tag{calligraphy} and \Tag{brush} are used in different genres, although both are handwriting-related.\par
One limitation of the current method is that its accuracy depends on the hyperparameters of the impression ensemble. As shown in Fig.~\ref{fig:font-impressions}, the current method sometimes discards reasonable impressions. This limitation comes from a difficult trade-off between the good ignorance of noisy labels and the necessary compensation for missing labels. Therefore, future work can be a more reliable impression ensemble by employing various machine learning-based strategies.\par
Another limitation of this paper is that we analyzed the correlations between the font impressions and the contexts (i.e., book genres in our experiments) and did not utilize the correlations to support actual typographic work. Nowadays, we can find machine learning-based typographic design generators. Future work will integrate our correlation analysis results into those generators as an empirical prior for realizing more sophisticated design outputs.\par
%
\par \vspace{3mm}
\noindent{\bf Acknowledgment}:\ This work was supported by JSPS KAKENHI Grant Number JP22H00540 and JST ACT-X Grant Number JPMJAX22AD and JST ACT-X Grant Number JPMJAX22AD.
%
%
\bibliographystyle{splncs04}
\bibliography{ref}
\end{document}